# Evaluation of Hindi to Punjabi Machine Translation System

**Vishal GOYAL and Gurpreet SINGH LEHAL**

**Department of Computer Science, Punjabi University**
**Patiala, India**
**{vishal, gslehal}@gmail.com**

## Abstract

Machine Translation in India is relatively young. The earliest efforts date from the late 80s and early 90s. The success of every system is judged from its evaluation experimental results. Number of machine translation systems has been started for development but to the best of author knowledge, no high quality system has been completed which can be used in real applications. Recently, Punjabi University, Patiala, India has developed Punjabi to Hindi Machine translation system with high accuracy of about 92%. Both the systems i.e. system under question and developed system are between same closely related languages. Thus, this paper presents the evaluation results of Hindi to Punjabi machine translation system. It makes sense to use same evaluation criteria as that of Punjabi to Hindi Punjabi Machine Translation System. After evaluation, the accuracy of the system is found to be about 95%.

## 1. Introduction

The present system involves Hindi as source language and Punjabi as target language. Both languages are closely related languages i.e. similar in respect to syntax, word order etc. Thus, ideal approach for translation process is direct approach. Every Machine translation undergoes an evaluation process for testing its accuracy to know its success. This paper will also explain the methodology adopted for evaluating the system and the results found after evaluation. The methodology followed for evaluation is same as that of Punjabi to Hindi Machine Translation system developed by Punjabi University Patiala. Both the systems are between the same languages, i.e., Hindi and Punjabi and reverse of each other. It is obvious choice to adapt the same methodology as that of already developed and tested system.

## 2. Evaluation Methodology

Based on the survey of existing evaluation methods for machine translation system and the evaluation criteria adopted by the developers of Punjabi to Hindi Machine Translation System, It is concluded the evaluation criteria adopted by latter system is suitable for the current system. Following are the steps that will be performed during evaluation:

1. Selection Set of Sentences: Test data will be selected.
2. Two type of subjective tests will be performed viz. Intelligibility and Accuracy.
3. Error test i.e. Word Error rate and Sentence Error rates will be performed.
4. Scoring Procedure for subjective tests will be devised.
5. Experimentation will be done using above tests on test data.
6. Analysis of the results from step 5 will be done.

The above steps will be discussed in detail in following sections of the paper.

### 2.1 Selection Set of Sentences:

Input sentences are selected from randomly selected news (sports, politics, world, regional, entertainment, travel etc.), articles (published by various writers, philosophers etc.), literature (stories by Prem Chand, Yashwant jain etc.), Official language for office letters (The Language Officially used on the files in Government offices) and blogs (Posted by general public in forums etc.). Simple as well as complex sentences of declarative, interrogative, imperative and exclamatory of varied length types have been included to test the system on every flavor. Following table show the test data set:

| | Daily News | Articles | Official Language Quotes | Blog | Literature |
|---|---|---|---|---|---|
| Total Documents | 100 | 50 | 01 | 50 | 20 |
| Total Sentences | 10000 | 3500 | 8595 | 3300 | 100450 |
| Total Words | 93400 | 21674 | 36431 | 15650 | 95580 |

**IJCSI**

Table 1: Test data set for the evaluation of Hindi to Punjabi Machine Translation System

## 2.2 Experiments

The survey was done by 50 People of different professions. 20 Persons were from Villages who only knows Punjabi Language and donot know Hindi and 30 persons were from different professions having knowledge of both Hindi and Punjabi Language. Average ratings for the sentences of the individual translations were then summed up (separately according to intelligibility and accuracy) to get the average scores. Percentage of accurate sentences and intelligent sentences is also calculated separately by counting down the number of sentences.

## 2.3 Intelligibility Evaluation

The evaluators do not have any clue about the source language i.e. Hindi Language. They judge each sentence (in target language i.e. Punjabi) on the basis of its comprehensibility. The target user is a layman who is interested only in the comprehensibility of translations. Intelligibility is effected by grammatical errors, miss-translations, and un-translated words.

## 2.3.1 Scoring

The scoring is done based on the degree of intelligibility and comprehensibility. A Four point scale is made in which highest point is assigned to those sentences that look perfectly alike the target language and lowest point is assigned to the sentence which is un-understandable. Detail is a follows:
*Score 3 :* The sentence is perfectly clear and intelligible. It is grammatical and reads like ordinary text.
*Score 2:* The sentence is generally clear and intelligible. Despite some inaccuracies, one can understand immediately what it means.
*Score 1:* The general idea is intelligible only after considerable study. The sentence contains grammatical errors &/or poor word choice.
*Score 0:* The sentence is unintelligible. Studying the meaning of the sentence is hopeless. Even allowing for context, one feels that guessing would be too unreliable.

## 2.3.2 Intelligibility Test Results

The response by the evaluators were analysed and following are the results:

- 70.3 % sentences got the score 3 i.e. they are perfectly clear and intelligible.
- 25.1 % sentences got the score 2 i.e. they are generally clear and intelligible.
- 3.5 % sentences got the score 1 i.e. they are hard to understand.
- 1.1 % sentences got the score 0 i.e. they are not understandable.

So we can say that about 95.40 % sentences are intelligible. These sentences are those which have score 2 or above. Thus, we can say that the direct approach can translate Hindi text to Punjabi Text with a tolerably good accuracy.

Table 2: Percentage Intelligibility of individual documents

|  | Daily News | Articles | Official Language Quotes | Blog | Literature |
|---|---|---|---|---|---|
| *% Intelligibility* | 99 | 90.5 | 90.7 | 90.8 | 87.4 |

## 2.3.3 Analysis

The main reason behind less accuracy for Literature documents is due to the language dialect used by the writer of the stories. Some writers use Rajasthani language, some uses Haryana dialect. Ans this resulted in less translation accuracy for this category. Otherwise for rest of the four categories, the quality of translation is better than other systems which will be discussed in following sections.

## 2.4 Accuracy Evaluation

The evaluators are provided with source text along with translated text. A highly intelligible output sentence need not be a correct translation of the source sentence. It is important to check whether the meaning of the source language sentence is preserved in the translation. This property is called accuracy.

## 2.4.1 Scoring:

The scoring is done based on the degree of intelligibility and comprehensibility. A Four point scale is made in which highest point is assigned to those sentences that look perfectly alike the target language and lowest point is assigned to the sentence which is un-understandable and unacceptable. The scale looks like:
*Score 3 :* Completely Faithful
*Score 2:* Fairly faithful: more than 50 % of the original information passes in the translation.

*Score 1:* Barely faithful: less than 50 % of the original information passes in the translation.
*Score 0:* Completely Unfaithful. Doesn't make sense.

### 2.4.2 Accuracy Test Results

Initially Null Hypothesis is assumed i.e. the system's performance is NULL. The author assumed that system is dumb and does not produce any valuable output. By the intelligibility of the analysis and Accuracy analysis, it has been proved wrong.

The overall score for accuracy of the translated text came out to be 2.63. The accuracy percentage for the system is found out to be 87.60%

Further investigations reveals that from 13.40% :

- 80.6 % sentences achieve a match between 50 to 99%
- 17.2 % of remaining sentences were marked with less than 50% match against the correct sentences.
- Only 2.2 % sentences are those which are found unfaithful.

A match of lower 50% does not mean that the sentences are not usable. After some post editing, they can fit properly in the translated text.

Table 3: Percentage Accuracy of individual documents:

|  | Daily News | Articles | Official Language Quotes | Blog | Literature |
|---|---|---|---|---|---|
| % Accuracy | 95 | 80.5 | 90.3 | 78.5 | 85.4 |

### 2.4.3 Analysis

The overall performance accuracy test of the system is quite good. But for Blog it is less than others. The reason is the use of slang which causes the failure of the translation software as the slang available in one language is not present in other language. Also un-standardized language cause more ambiguities.

### 2.5 Error Analysis

To check the Error rate of the Direct Translation System, some quantitative metrics are also evaluated. These include:

- *Word Error Rate:* It is defined as percentage of words which are to be inserted, deleted, or replaced in the translation in order to obtain the sentence of reference.
- *Sentence Error Rate:* It is defined as percentage of sentences, whose translations have not

matched in an exact manner with those of reference

Error analysis is done against pre classified error list. All the errors in translated text were identified and their frequencies were noted. Errors were just counted and not weighted. Main categories of errors are:

*A.* There are some words in Hindi that can be translated into different forms but the meaning is almost same and their translation depends upon grammatical context. For

Example : word **सजा (decorate)**

| Input | : | उसने सारा घर सजा दिया |
|---|---|---|
| **Output** | : | ਉਸਨੇ ਸਾਰਾ ਘਰ ਸੱਜਿਆ ਦਿੱਤਾ |
| **Input** | : | उसने सजा हुआ घर देखा |
| **Output** | : | ਉਸਨੇ ਸੱਜਿਆ ਹੋਇਆ ਘਰ ਵੇਖਿਆ |

In the above examples, the word सजा can be translated as decorated or decorate. Similarly, word हो can be translated as ਹੋ or ਹੋਵੇ

*B.* Hindi Word और (And) can be translated as ਅਤੇ (And) and ਹੋਰ (More/ Another) . Example : word और (And/ More/ Another)

**Input :** उनके और पाइंट के विद्यार्थियों के बीच का संवाद बेहद रोचक रहा।

**Output :** ਉਨ੍ਹਾਂ ਦੇ **ਹੋਰ** ਪਾਇੰਟ ਦੇ ਵਿਦਿਆਰਥੀਆਂ ਦੇ ਵਿੱਚ ਦਾ ਸੰਵਾਦ ਬੇਹੱਦ ਰੋਚਕ ਰਿਹਾ ।

**Input :** राजस्थान की शुरुआत बेहद खराब रही और एक बार दबाव में आने के बाद उसके सभी बल्लेबाज अपना विकेट फेंककर चलते बने।

**Output :** ਰਾਜਸਥਾਨ ਦੀ ਸ਼ੁਰੁਆਤ ਬੇਹੱਦ ਖ਼ਰਾਬ ਰਹੀ ਅਤੇ ਇੱਕ ਵਾਰ ਦਬਾਅ ਵਿੱਚ ਆਉਣ ਦੇ ਬਾਅਦ ਉਸਦੇ ਸਾਰੇ ਬੱਲੇਬਾਜ ਆਪਣਾ ਵਿਕੇਟ ਸੁੱਟਕੇ ਚਲਦੇ ਬਣੇ ।

### 2.5.1 Word Error Analysis

After robust analysis of Word Error rate is found out to be 5.2% Which is comparably lower than that of general systems, where it ranges from 9.5 to 12%.

Table 4: Percentage type of errors out of the errors found

| Wrongly translated word or expression | 10.3% |
|---|---|
| Addition or removal of words | 6.7% |
| Untranslated words | 15.5% |
| Wrong choice of words | 67.5% |

From the above table, it is concluded that majority of the errors are due to wrong choice of words, means the WSD module of the system must be improved. Further, the bilingual dictionary improvements can reduce the wrongly translated and untranslated words errors.

Table 5: Word Error rate Percentage

|  | Daily News | Articles | Official Language Quotes | Blog | Literature |
|---|---|---|---|---|---|
| WER % age | 3.1 | 4.4 | 4.7 | 5.2 | 5.2 |

2.5.2 Sentence Error Rate Percetage:

The Sentence error rate comes out to be 42.4%

|  | Daily News | Articles | Official Language Quotes | Blog | Literature |
|---|---|---|---|---|---|
| SER % age | 15.4% | 25.2% | 20.7% | 40.68% | 42.14% |

2.5.3 Analysis:

As discussed earlier, the WER and SER of un-standardized matter i.e. Blog and Literature is higher than the standardized matter. It strengthens the fact that better input gives the better output. If some pre editing of the text is performed then better results may be expected.

## 3.0 Comparison with other existing systems

| MT SYSTEM | Accuracy |
|---|---|
| RUSLAN | 40% correct 40% with minor errors. 20% with major error. |
| CESILKO (Czech-to-Slovak) | 90% |
| Czech-to-Polish | 71.4% |
| Czech-to-Lithuanian | 69% |
| Punjabi-to-Hindi | 92% |
| Hindi-to-Punjabi | 95.12% |

From the above table, it is clear that the system is outperforming in comparison to others. Thus system is anonymously acceptable to practical use.

## 4.0 Conclusion

From the above analysis, it is concluded the overall accuracy of Hindi to Punjabi machine translation system is found to be 95.12%. The accuracy can be improved by improving and extending the bilingual dictionary. Even robust pre processing and post processing of the system can improve the system to greater extent. This system is comparable with other existing system and its accuracy is better than those.